\documentclass[letterpaper]{article} 
\usepackage[]{aaai24}  
\usepackage{times}  
\usepackage{helvet}  
\usepackage{courier}  
\usepackage[hyphens]{url}  
\usepackage{graphicx} 
\usepackage{natbib}  
\usepackage{caption} 
\usepackage{algorithm}
\usepackage{algorithmic}

\usepackage{newfloat}
\usepackage{listings}
\DeclareCaptionStyle{ruled}{labelfont=normalfont,labelsep=colon,strut=off} 
\floatstyle{ruled}
\newfloat{listing}{tb}{lst}{}
\floatname{listing}{Listing}
\nocopyright

\title{For those who don't know (how) to ask:\\
Building a dataset of technology questions for digital newcomers}

\author{
    Evan Lucas, Kelly S. Steelman, Leo C. Ureel, Charles Wallace
}
\affiliations{
    Michigan Technological University\\

    1400 Townsend Drive,
    Houghton MI 49931-2122 USA\\
    eglucas@mtu.edu, ureel@mtu.edu, steelman@mtu.edu, wallace@mtu.edu
   }

\begin{document}

\maketitle

\begin{abstract}
While the rise of large language models (LLMs) has created rich new opportunities to learn about digital technology, many on the margins of this technology struggle to gain and maintain competency due to lexical or conceptual barriers that prevent them from asking appropriate questions. Although there have been many efforts to understand factuality of LLM-created content and ability of LLMs to answer questions, it is not well understood how unclear or nonstandard language queries affect the model outputs. We propose the creation of a dataset that captures questions of digital newcomers and outsiders, utilizing data we have compiled from a decade's worth of one-on-one  tutoring. In this paper we lay out our planned efforts and some potential uses of this dataset.
\end{abstract}

\section{Introduction}
\label{sec:intro}

\begin{quote}
    {\em And [regarding] the one who doesn't know to ask, you will open [the conversation] for him.

    Pesach Haggadah, Magid, The Four Sons 5}
\end{quote}
    
The Haggadah, the Jewish text that provides guidance in instructing young people about the story of Exodus through the Passover Seder, makes special mention of those who, for a variety of reasons --- including lack of confidence or context --- are unable to formulate questions that can initiate their understanding. Newcomers to digital technology face similar barriers that can halt or limit their understanding: unfamiliarity with technical terminology; unawareness of relevant contextual details; anxiety that can erode self efficacy and cognitive load. Even when tutors and mentors are available, such barriers can inhibit the productive dialogue that builds true digital competence. 

For over a decade, members of our project team have assisted those on the margins of digital technology through the BASIC (Building Adult Skills in Computing) program, helping them strengthen higher order skills and increase self-efficacy \citep{steelman2021breaking,atkinson2016breaking,steelman2017eliciting}.
Our tutors employ a variety of rhetorical strategies to understand learners’ mental models, demonstrate effective problem solving behavior, assuage tech anxiety
\citep{steelman2017identifying}, 
and work with learners to craft solutions. These strategies have been used in one-on-one tutoring sessions, primarily at local public libraries. Advances in Large Language Models (LLMs) and chat technology suggest exciting ways to provide assistance when human tutors are not available. These advances, however, do not entirely address the complexities of our tutoring sessions, where learners often do not know what they need nor what questions to ask.


The overall aim of this project is an exploration of applying pedagogically motivated rhetorical strategies developed for interactive personal tutoring to LLM chatbots to create a dialogical automated tutoring system. We also seek to ascertain to what degree a dialogue-based automated learning chat system could be used to help learners assess and build digital competencies. We would like to incorporate these systems into existing digital literacy assistance and instruction currently in place at public libraries in a way that best serves learners and staff. We are working with 
the Superiorland Library Collective, a network of libraries across Michigan's rural Upper Peninsula, 
to get context and experience from professionals currently engaged in this work.

\section{Proposed Datasets}
\label{datasets}

We plan to develop three datasets. The first will come from mining existing online forums on computer-related topics; the second will tap into our extensive records of tutor-learner interaction from the BASIC program; and the third will be developed from interactions at our partners in the Superiorland Library Collective. In addition to our own use, these datasets will be made publicly available to help further research into digital literacy and educational chatbots. 

\subsection{Mining existing datasets}
\label{yahoo}
The first steps in developing this dataset will be to mine existing datasets, including the Yahoo Answers dataset \cite{zhang2015character},  with question and answers relevant to digital literacy. The advantage of Yahoo Answers is that the \emph{Computers \& Internet} class contains 140,000 training samples and 6,000 testing samples to review as potential source material. Some or all of this subset will be human-labeled for topic and whether or not the question demonstrates digital literacy relevant to the question being asked. Each question will also be coded based on the question characteristics (e.g., specific shortcomings like missing information or imprecise language).

\subsection{Digital literacy tutor input}
\label{tutor_interviews}

Although the Yahoo database provides a rich source of learner-generated questions, it does not include information about the ancillary activities that tutors engage in to clarify learners' needs, understand the context better, and take advantage of teachable moments. BASIC tutors regularly log the details of their tutoring sessions; after a decade of these sessions, we have compiled a database with hundreds of rich records of tutor-learner interactions, articulating common problems, frustrations, and strategies. To build a dataset with details about specific knowledge gaps observed in past tutoring, we will review this corpus of tutor notes, conduct interviews with experienced digital literacy tutors, local information technology experts, and other educators. Interviewees will be prompted with representative questions containing shortcomings in digital literacy, and asked to discuss why the question is improperly framed and generate a suggested strategy for probing the learner for more information and answering the question in a way that is tuned for the learner's current knowledge and experience level.

\subsection{Recorded and annotated tutor-learner interaction}
\label{ongoing_experiment_data}
The third dataset will be created from recordings of actual tutoring sessions. Project personnel will visit five Superiorland Library Collective member libraries across Michigan's Upper Peninsula 
and engage in tutoring sessions with local learners.  These interactions will be recorded and annotated by tutors to create a useful dataset for future efforts in this area. The detail given to annotating individual sessions will allow us to include information about the learners understanding of the topic, including any shortcomings that directly impact their ability to form questions, and the answers that they are actually seeking.

\section{Ensuring factuality with unclear questions}
\textbf{RQ1: How can LLMs detect poor questions and provide effective probes and clarifying questions to improve the quality and relevance of its responses?}

Determining factuality of LLM outputs \cite{wang2023survey} has major implications for educational chatbots, where an unclear question may cause the LLM to hallucinate content that is incorrect or misguiding, or answer an entirely different question. A corpus of unclear questions along with tutor-generated probes and corrected questions and responses would enable research into the impact of question wording on factuality of generated responses.

Training a classifier to determine the level of competency implied by the question would be useful for creating a model of the learner and tailoring responses. If such a classifier can be created with good accuracy on a relatively small dataset, it may allow the mining of larger datasets to help generate additional materials for this project. Additionally, a classifier may enable the system to assess the level of scaffolding that the learner needs to support problem solving and when that scaffolding can be reduced or removed.

\section{Question classification}
\textbf{RQ2: How can pedagogically motivated rhetorical strategies and tactics from interactive personal tutoring be integrated with the chat technology of large language models to create a dialogical automated tutoring system?}


A key component of our tutoring process is ascertaining the digital competency of the learner. Many taxonomies and inventories for digital competency exist \cite{boot2015computerproficiency,roque2018mobiledevice,roque2021wirelessnetwork,vanlaar201821stcentury,vuorikari2022digitalcompetence}, but they are typically used in a simple, non-interactive survey format for assessment. We will explore how to incorporate these instruments into a dynamic, dialogical setting that could be implemented by an LLM. Ideally, such a system would be able to identify the information the learner is missing in their question and be able to educate them or share resources to help improve their understanding. Our proposed system must itself generate effective questions \citep{elkins2023useful} that probe a learner’s digital competency, then move toward a solution using discourse appropriate to their competency level.

\section{Limitations and expected challenges}
Much like the learner questions we are documenting, there is some uncertainty around the best means to accomplish our tasks. We anticipate challenges in ensuring diversity of the question topics and determining how best to utilize this dataset for generation.

We have identified the potential for sampling bias in our proposed data sources. First, the interactions from Yahoo Answers involve people who are already online and therefore have a level of technological familiarity beyond some of our target audience. Similarly, our in-person interactions come from a learner demographic with the means and resolve to show up and ask questions. Finally, our tutors are far removed from being in the position of not understanding core concepts necessary to ask questions. By combining these different sources and using them to prompt additional questions from the tutor group, we hope to mitigate some of this sampling bias. We also have identified the possibility for overfitting our question classification model to detect aspects of the data, such as specific language used by a learner or misspellings only present in the Yahoo Answers data. 

In order to train the model to learn tutors' dialogical flow, the training data must include full transcripts engineered into a chat format. As our tutoring notes and Yahoo Answers formats are not in dialog format, text data augmentation will be required. Care must be taken in this process to ensure that resulting dialog accurately represents authentic tutor-learner interactions.

\section{Conclusion}
In this short paper, we lay out our proposal for creating a new dataset of questions asked by digital newcomers and outsiders. This dataset will exemplify questions that are asked using unclear or incorrect vocabulary. We also briefly discuss some possible research directions that we are considering that would utilize this dataset.

\bibliography{references.bib}

\appendix



\end{document}